# TEXT-TO-FACE GENERATION WITH STYLEGAN2

D. M. A. Ayanthi and Sarasi Munasinghe

Department of Computer Science, Faculty of Science,
University of Ruhuna, Wellamadama, Matara, Sri Lanka

## ABSTRACT

*Synthesizing images from text descriptions has become an active research area with the advent of Generative Adversarial Networks. The main goal here is to generate photo-realistic images that are aligned with the input descriptions. Text-to-Face generation(T2F) is a sub-domain of Text-to-Image generation(T2I) that is more challenging due to the complexity and variation of facial attributes. It has a number of applications mainly in the domain of public safety. Even though several models are available for T2F, there is still the need to improve the image quality and the semantic alignment. In this research, we propose a novel framework, to generate facial images that are well-aligned with the input descriptions. Our framework utilizes the high-resolution face generator, StyleGAN2, and explores the possibility of using it in T2F. Here, we embed text in the input latent space of StyleGAN2 using BERT embeddings and oversee the generation of facial images using text descriptions. We trained our framework on attribute-based descriptions to generate images of 1024x1024 in resolution. The images generated exhibit a 57% similarity to the ground truth images, with a face semantic distance of 0.92, outperforming state-of-the-artwork. The generated images have a FID score of 118.097 and the experimental results show that our model generates promising images.*

## 1. INTRODUCTION

With the advent of Generative Adversarial Networks (GANs)[1], image generation has achieved ground-breaking results because of its ability to generate high-quality images that show a close resemblance to real images. However, the original GAN did not have the ability to control the images that it was trained to generate and render images that meet a given specification. In order to overcome this potential issue, various conditional GAN models were proposed over time for different tasks. Text-to-Image generation (TTI) is one such application. It refers to the generation of an image that meets the context specified in a natural language description. We can also call this as the inverse of image captioning as it tries to learn a mapping from the text space to the image space. This emerging research topic is less explored mainly because of its complexity and challenging nature. However, it has a huge number of interesting applications including image processing tasks like image editing, computer-aided design, and computer game development.

Table 1. Samples from Text2FaceGAN dataset.

| 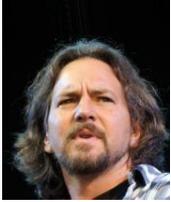 | 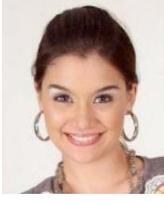 |
|---|---|
| The man has a chubby face. He sports a goatee with sideburns. His hair is black in color. He has narrow eyes and a slightly open mouth. The man looks young. | The woman has oval face and high cheekbones. She has big lips with arched eyebrows and a slightly open mouth. The smiling, young attractive woman has heavy makeup. She's wearing earrings, necklace and lipstick. |

The traditional methods for TTI, mostly use similar frameworks consisting of a pretrained text encoder to encode the text descriptions to a semantic vector and a conditional GAN model as the image decoder. These setups were mostly limited to simpler images like birds and flowers due to the complexity of learning the mapping between the text and the image spaces and the availability of datasets like the CUB bird dataset [2], Oxford Flower bird dataset [3], and MS-COCO dataset [4].

Text-to-Face generation (TTF) is a sub-topic coming under TTI which is particularly focused on generating human faces from natural language descriptions. Similar to TTI, TTF has two main targets; 1) to generate realistic, high-resolution facial images, 2) to generate images that are well aligned with the input descriptions. Compared to TTI, TTF has more value considering the public safety domain. This can be used mainly in criminal investigation and in the preparation of datasets for bio-metric research involving face data like face recognition and age estimation. Realistic faces can be generated to assist identifying criminals by automating the task of a sketch artist and in place of a sketch, this could be used to generate an image that is favourable in the investigation. The face datasets that are used in research like face detection and age estimation are typically compiled by scraping images from the internet, mostly, without the consent of the owner and this raises ethical concerns. Another major issue with these datasets is racial biases like minorities not being represented properly. Using datasets composed of synthetically generated images is a possible solution to this.

Most of the existing T2F frameworks have only been able to produce low-resolution images that have poor consistency with the input descriptions. It is important to be able to generate high-resolution images to use TTF in the above-mentioned applications. Recent progress on GANs has established a remarkable paradigm on image generation in terms of quality, fidelity, and realism. StyleGAN2[5] is one of the most significant GAN frameworks that has been introduced and with its style-based generator architecture it is able to produce high-resolution images with unmatched photorealism. It has not only been trained to generate human faces but also other images. However, when comparing the images generated through TTF frameworks and those generated through GAN models specialized for face generation, like the StyleGAN2 there is a clear difference. The traditional multi-staged architectures and the progressive training of TTF frameworks have not been able to generate quality images. Also, it can be seen that the prospect of using a high-resolution generator for TTF has not been considered. Taking this into consideration we propose a novel framework for TTF using the StyleGAN2 generator. Here we aim to represent text descriptions in the latent space of the StyleGAN2 generator and thereby generate facial images. We propose a simpler model that can find a mapping between the text space and the image space and use this as a means to generate images from natural language descriptions. We use BERT [6] sentence encoder as the language model and the StyleGAN2 generator as the image generator. We use the sentence level embeddings obtained from the sentence encoder to learn a text-to-latent model, that maps the descriptions to the input space of the state-of-the-art generator, StyleGAN2 to generate images.

Our main contributions are as follows,

- Propose a novel TTF framework using StyleGAN2 as the image generator and BERT sentence encoder as the language model
- Generate high-resolution images that are properly aligned with the input descriptions.

The rest of the paper is organized as follows. In Section 2, we discuss the related work and in Section 3 we discuss the proposed methodology including the datasets and the other preliminary frameworks used in this work. The experimental analysis, results obtained and the evaluation is presented in Section 4 and lastly, in Section 5 we discuss the conclusion and the future work.

## 2. RELATED WORK

Text-to-Face generation has been greatly influenced by the work done in Text-to-Image generation. Apart from that, image generation (Face generation) is another domain to be considered. This section discusses the important work that was done in these three fields.

### 2.1. Text-to-Image generation

The main goals of these models are two-fold. One is to generate high-quality images and the other is to generate images that are properly aligned with the input description. During the early stage of the development of these models, it was focused on generating high-quality images. The first model to take advantage of a GAN was presented in 2016 by Reed et al [7]. They contributed with two models for text-to-image generation using conditional GANs. They used a pretrained Char-CNN-RNN network [8] as the text encoder, a model similar to the DCGAN [9] as the image decoder, and produced images of 64x64 and 128x128. This model was unable to find a good mapping between the keywords and the image features due to the direct concatenation of the text embeddings with noise inputs. To overcome this issue, StackGAN [10] was proposed. It was a GAN model with two stages. The first stage captures some information in the description and generates an initial, low-resolution image. During the second stage, the image is refined with the description to produce images of 256x256. This network, with hierarchically nested generators, is used in most of the later approaches as well [10][11][12]. Even so, generating high-resolution images like 1024x1024 is very expensive using this kind of architecture.

When it was possible to generate realistic images, the next target was to generate images to improve the similarity between the generated image and the input description. Among the various proposed approaches, the attention mechanism is significant. This was first proposed in the context of image generation by Xan in AttnGAN. They introduced a word-level attention mechanism that enabled the generation of fine-grained images of 256x256. However, the word level attention alone was not enough and lead to the generation of unrealistic images. Another issue with the model was that as this model was trained on descriptions with mostly one sentence, it was unable to handle longer sentences. In spite of these shortcomings, this attention mechanism has been the base model for most of the work that was conducted later [13][14].

## 2.2. Image generation (Face generation)

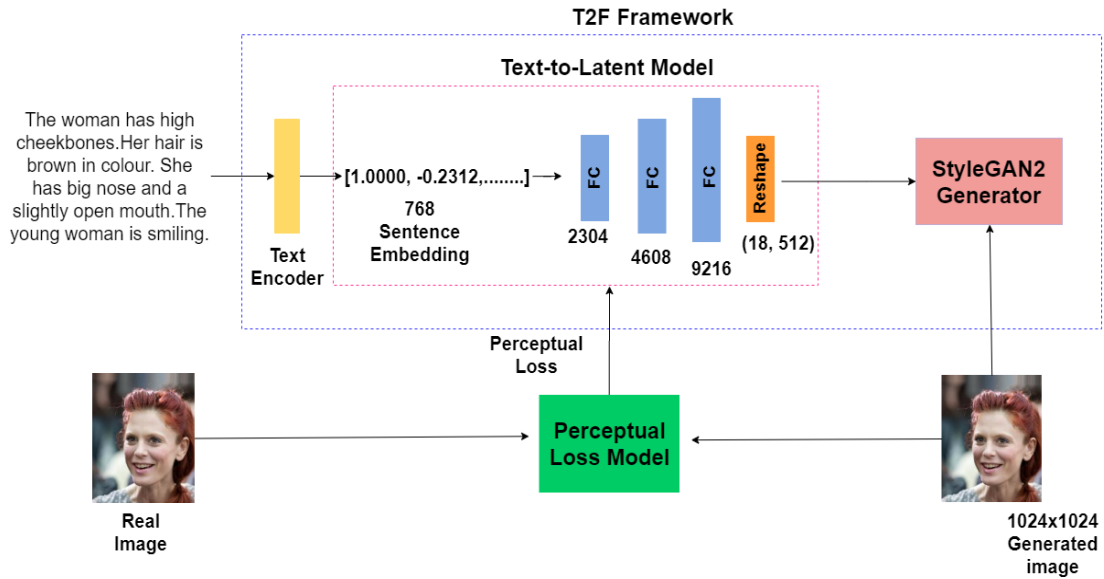

Figure 1. Proposed Model Architecture

Image generation models using GANS, learn a mapping from a noise vector to the normal distribution corresponding to real images. With the availability of two large datasets, LFW [15] and CelebA[16], face synthesis is also a trending research area. Most of these models are based on conditional GANs. Some of the popular models are ProGAN [17], StyleGAN [18], and StyleGAN2. Instead of attempting to train all the layers of the generator and the discriminator at once, in ProGAN they have gradually grown the GAN one layer at a time, to generate high-resolution images gradually. Using this approach, they have been able to start from images of 4x4 and gradually increase up to 1024x1024. StyleGAN and StyleGAN2 are also built on top of ProGAN, but they have control over the style factors, unlike ProGAN. The images generated by StyleGAN2 are very realistic that they cannot be easily recognized even by humans as fake, generated images. The only disfunction here is that there is no control over the generated images. Therefore, it cannot generate a particular face at our request.

## 2.3. Text-to-Face Generation

Compared to the work done in T2I there is a far lesser amount of work done in T2F. The main reason behind this is the variety of facial attributes in terms of ethnicity, age, and the facial descriptions being vague about the attributes. Even with the said challenges, there is a smaller number of inspiring work done in text-to-face generation. In the project T2F by Akanimax [19], he proposed to encode text descriptions into a summary vector using an LSTM and use ProGAN as the generator. As the image quality was poor, they used MSG-GAN [20] as the generator and improved the image quality. Text2FaceGAN [21] was based on the GAN-INT-CLS architecture by Reed et al [7]. The Text2Face dataset was also introduced using the attributes of the CelebA dataset and an algorithm for caption generation. They could only produce images of resolution 64x64. Here they also showed howInception Score [22] which is a generally used metric in GAN evaluation is not suitable to be used with facial images.

FTGAN [23], proposed an architecture that combined the training of the text encoder and the image decoder that was done separately so far. The main idea was that to generate quality images the text encoder and the image decoder needs to be trained together because when using a pre-trained text encoder, the input to the image decoder is too dependent on the output of the text encoder. However, with this proposed architecture, FTGAN produced images up to 256x256. Another similar approach is given in [24]. Instead of the BiLSTM in FTGAN here they are using a Char-CNN-RNN [8] to obtain the semantic vector and train both the text encoder and the image decoder at the same time. With this approach, they obtained slightly better images but still, they could produce images only up to 256x256. TTF-HD [25] proposed a framework consisting of a multi-label text classifier, an image label encoder, and an image generator to generate facial images of 1024x1024. With the multi-label classifier, they have been able to consider a feature disentangled latent space and focus on the diversity of the images generated other than the quality and the semantic consistency. However, utilizing the ability of StyleGAN2 to produce high-quality facial images has not been considered much in this work.

When referring to the relevant literature we could observe that most T2F models were based on architectures introduced for T2I generation and yet the image quality was far less than those naïve GAN models specialized for face generation. Generating high-resolution facial images is still a problem that has not been addressed properly. Also, the usage of existing high-resolution generators in T2F is minimal. Therefore, there is a need to explore the possibility of using face generation models like StyleGAN2 in T2F to generate high-resolution images. It was also identified that newer language models like BERT have not been used for this task. Based on these gaps, we conduct experiments in this paper to determine the possibility of using StyleGAN2 in T2F and develop a novel framework for T2F to generate high-resolution facial images that are consistent with the input descriptions.

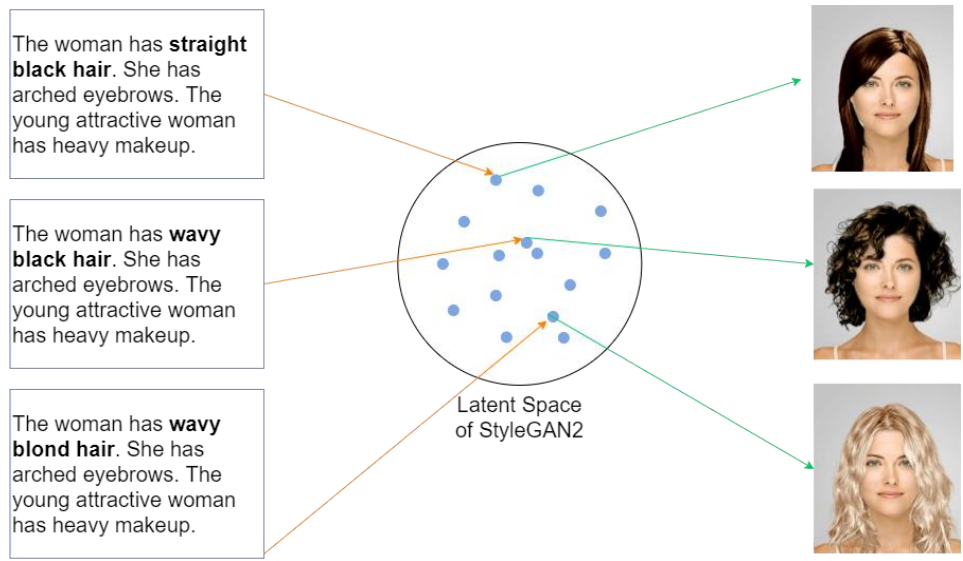

Figure 2. Representation of a description in the latent space of the StyleGAN2 and the corresponding images when attributes in the description are changed.

## 3. PROPOSED METHODOLOGY

This section provides the details of the dataset, proposed architecture for T2F and the evaluation metrics. The proposed approach is a pipeline consisting of several steps and the complete architecture is presented in Figure 1.

## 3.1. Dataset

There are several datasets used by T2F models like the Face2Text[26], SCU-Text2Face[23], and Text2FaceGAN[21]. Face2Text dataset consists only of 400 image-description pairs, which is insufficient and the SCU-Text2Face dataset is not publicly available. Therefore, we used the Text2FaceGAN dataset. This is a set of captions generated for the images in the CelebA dataset, using the attribute list provided. The CelebA dataset has 40 facial attributes and the Text2FaceGAN dataset uses them in the captions that have been created using an algorithm. Table 1 shows some sample records from the dataset. We chose a sample of 6000 images from the Text2FaceGAN dataset for training and testing the model due to the limited resources.

| The woman has oval face. She has straight hair which is black in colour with bangs. She has big lips. The young attractive woman has pale skin and heavy makeup. She's wearing lipstick. | | | | The man sports a 5 o'clock shadow. He has straight hair which is brown in colour with bangs. He has pointy nose. The man looks young. | | | |
|---|---|---|---|---|---|---|---|
| Real Image | Experiment 01 | Experiment 02 | Experiment 03 | Real Image | Experiment 01 | Experiment 02 | Experiment 03 |
| 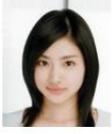 | 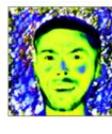 | 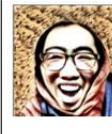 | 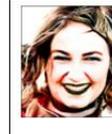 | 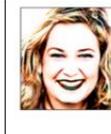 | 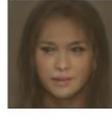 | 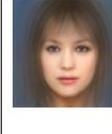 | 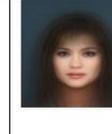 |
| | Experiment 04 | Experiment 05 | Experiment 06 | | Experiment 04 | Experiment 05 | Experiment 06 |
| | 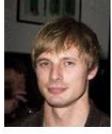 | 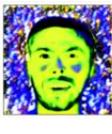 | 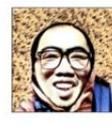 | | 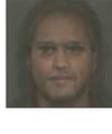 | 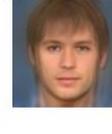 | 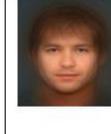 |

Figure 3. Experimental results obtained for all experiments conducted in both Z and W latent spaces.

## 3.2. Proposed Model Architecture

The proposed architecture for T2F consists of four modules. 1) Text encoder 2) Text-to-Latent model 3) Image generator 4) Perceptual loss model. In the following sections we have discussed them in detail.

### 3.2.1. Text encoder

The text encoder is used to extract the semantic vectors of the input descriptions. This is done to bring the text space and the image space to some common grounds, to learn how to map the descriptions to the images. We use the state-of-the-art language model, BERT (Bidirectional Encoder Representations from Transformers) as the text encoder to extract the semantic vectors from the input descriptions in the form of a sentence embedding. We used the BERT-as-a-service facility to ease the task.

BERT model producing state-of-the-art results in major Natural Language Processing tasks was the main reason for selecting BERT as our language encoder. The technical innovation behind BERT is the application of bi-directional training of transformers into language processing tasks that gives the model a deeper sense of the language context and flow than single-directional models. So, unlike the other context-free word embeddings like GloVe[27] or Word2Vec[28], BERT embeddings are produced considering the context the words are used in. This was another reason for the selection of BERT as the language encoder. We chose the BERT BASE model, which has a stack of 12 transformer layers and returns an embedding vector of 768 dimensions for a given description.

### 3.2.2.   Text-to-Latent model

To use text as an input to the StyleGAN2, we need to learn how to represent text in the latent space of the StyleGAN2 model. For this purpose, we build a model that maps the input text description in the form of a text embedding to the input latent space and uses this as the input to the StyleGAN2 generator. This in turn allows us to control the images generated through the StyleGAN2 using a text description. Figure 2. gives an idea of the representation of descriptions in the latent space.

In StyleGAN2 there are two latent spaces; the initial latent space Z and the intermediate latent space W. These latent spaces have different properties and a lot of experiments are carried out to identify the properties and get a good understanding of these latent spaces at present. In representing the text in the latent space of the StyleGAN2 generator using the text-to-latent model, we experimented on both the latent spaces.  The proposed text-to-latent space model is a multi-layer perceptron consisting of fully connected layers to transform from the BERT distribution space to the input latent space of the StyleGAN2.

### 3.2.3.   StyleGAN2 generator

The output of the text-to-latent model will be passed to the high-resolution generator, StyleGAN2. It is an improved version of StyleGAN and was released by NVIDIA in 2020. This high-resolution generator produces faces with unmatched realism. Therefore, in this paper, we explore ways to represent text in the input latent space of the generator and use it in T2F rather than training a new generator from scratch. The main reason for the selection of the StyleGAN2 generator is the state-of-the-art results it produced in face generation with the introduction of features like weight demodulation, path length regularization, and removal of progressive growing. These features have led the way to overcome problems in the StyleGAN architecture like the generation of phase artifacts and the droplet effect.

### 3.2.4.   Perceptual Loss Model

When coming to image enhancement work like image super-resolution, colorization and style transfer the loss function intends to evaluate how far the generated/predicted output of the model is from target/ground truth image, to train the model to minimize the loss. The goal of our study is to define a model to represent text in the latent space of StyleGAN2 to control the images generated.  In this case, as well, we need to visually match the generated image with the ground truth i.e., the features generated in the image should be close to the features in the real image. The most commonly used loss functions in image enhancement processes are the pixel loss based on mean squared error (MSE), root mean squared error (RMSE), or peak-signal-to-noise ratio (PSNR). However, we chose feature loss or perceptual loss[29], which is a better measure because perceptual loss allows to reconstruct finer details of images compared to per-pixel loss. We used VGG16[30] to extract features of the input images from selected layers for the perceptual loss calculation.

The generated image from the StyleGAN2 generator and the corresponding real image are fed into the perceptual loss model. The activation maps, called feature maps, capture the result of applying the filters to input images. The feature maps close to the input detect small or fine-grained detail, whereas feature maps close to the output of the model capture more general features. Our aim is to generate images representing features of the ground-truth image.

Therefore, we need a balance of layers from the top and bottom. We experimented with several layer combinations for this reason. The difference in the selected feature maps of the real image and the generated images were used to update the text-to-latent model.

### 3.3. Evaluation Metrics

The goal of our model is to generate realistic images that are closely aligned with the input descriptions. The alignment of the images is measured by comparing the distance between the facial features of both images and the cosine similarity of them. The distance between the features is called Face Semantic Distance (FSD) and the similarity is called Face Semantic Similarity (FSS). FSD and FSS are calculated using equations (1) and (2).

$$Face\ Semantic\ Distance = \frac{1}{N}\sum_{i=0}^{N}\left|\left(F_{G_i}\right) - \left(F_{GT_i}\right)\right| \qquad (1)$$

$$Face\ Semantic\ Similarity = \frac{1}{N}\sum_{i=0}^{N}cos\left(F_{G_i} - F_{GT_i}\right) \qquad (2)$$

In the above equations, $F_{G_i}$ is the feature vector of the generated image and $F_{GT_i}$ is the feature vector of the real image. Cos indicates the cosine similarity between the feature vectors.

Generated images will be compared against the real images to measure how far they are realistic using the FID score[31]. It summarizes how similar the real and generated images are in terms of statistics on computer vision features of the raw images calculated using the Inceptionv3 model [32] used for image classification. Lower scores indicate the two groups of images are more similar or have more similar statistics. Fréchet distance also called the Wasserstein-2 distance is calculated using the equation (3).

$$d^2 = \left||mu_1 - mu_2\right||^2 + Tr\left(C_1 + C_2 - 2\sqrt{C_1 * C_2}\right) \qquad (3)$$

The score is referred to as d2, showing that it is a distance and has squared units. The "mu1" and "mu2" refer to the feature-wise mean of the real and generated images. The C1 and C2 are the covariance matrix for the real and generated feature vectors, often referred to as sigma. The ||mu1 − mu2||2 refers to the sum squared difference between the two mean vectors. Tr refers to the trace linear algebra operation (the sum of the elements along the main diagonal of the square matrix). The sqrt is the square root of the square matrix, given as the product between the two covariance matrices.

All images were scaled to 299x299 before calculating the scores.

Table 2. Summary of all the experiments conducted.

| Experiments in the initial latent space Z | | Experiments in the intermediate latent space W | |
|---|---|---|---|
| Experiment No. | Layers of VGG16 | Experiment No. | Layers of VGG16 |
| 01 | Conv4_3 Conv5_3 | 04 | Conv4_3 Conv5_3 |
| 02 | Conv3_2 Conv4_2 Conv5_2 | 05 | Conv3_3 Conv4_3 Conv5_3 |

| 03 | Conv3_2<br>Conv4_2<br>Conv5_2 with Hyper-columns | 06 | Conv1_2<br>Conv2_2<br>Conv3_2<br>Conv4_3 |
|----|--------------------------------------------------|----|-------------------------------------------|

## 4. EXPERIMENTS AND EVALUATION

This section discusses the extensive experimental analysis that has been carried out to evaluate the performance of the proposed model.

In this paper, we aim to represent text in the input latent space for StyleGAN2 to produce quality images that can be controlled using a text description. Therefore, the latent space of the StyleGAN2 generator is of great importance. The architecture of the StyleGAN2 generator uses two latent spaces; Z and W. The initial latent space Z is an entangled latent space and the intermediate latent space W shows more properties of disentanglement. However, the dimensionality of both the latent spaces is 512. In developing the proposed framework to project text to the latent space of the StyleGAN2 model, first, we need to determine the latent space that works best for our framework. The optimization of the Text2LatentSpace model is done using the perceptual loss model that uses a perceptual loss function between the feature maps of the generated images and the real images obtained through the VGG16 network. Different layers in the VGG16 network extract different features. Therefore, it is necessary to choose a suitable layer or layer combination to be used as the feature extractors.

In this context, there were two sets of experiments designed to reach our objectives. One to choose the most suitable latent space and the other to choose the best layer combination for feature extraction in the perceptual loss model. Table 2 gives a summary of all the experiments conducted with different configurations. The layers here are named following the layer names used in the VGG16[30]. Figure 3. shows results obtained from each experiment.

Through the experimental analysis, we observed that experiments conducted in the initial latent space were not successful. The model was unable to recognize the features in the descriptions and led to the generation of images with very slight changes that did not correspond to the descriptions. Apart from that, the generated images were unrealistic. This was due to the entangled nature of the initial latent space. However, the experiments conducted by projecting the descriptions to the intermediate latent space of the StyleGAN2 generator produced better results. These images were aligned with the descriptions as well as more realistic. From these experiments, we observed that experiment 05, which was done in the intermediate latent space using the conv3_3, conv4_3, and conv5_3 produced the best results both in terms of realism and semantic alignment. This model was trained for 500 epochs with an initial learning rate of 0.0001 and the Adam optimizer. Table 3 shows some images generated with their descriptions.

### 4.1. Qualitative Evaluation

**Image quality:** Generating visually appealing facial images is one of the main goals of this paper. Figure 4. shows some example facial images generated with the proposed model. We can see that all facial features have been correctly rendered in the generated images and the results are visually appealing to a greater extent. It also shows the ability of the proposed model to generate various faces across different facial features like gender, hair, smile, and age.

Table 3. Sample generated images and their input descriptions.

| | | | |
|---|---|---|---|
| The man has a double chined face. He sports a 5 o'clock shadow. He has a receding hairline. He has big lips and big pointy nose and a slightly open mouth. The man is smiling. He's wearing necktie. | 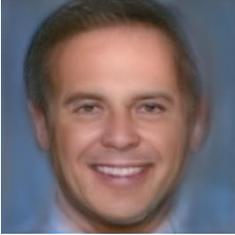 | The woman has oval face and high cheekbones. She has wavy hair which is brown in colour with bangs. She has big lips and pointy nose with arched eyebrows. The young attractive woman has heavy makeup. She's wearing a necklace and lipstick. | 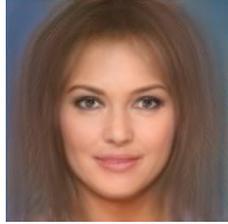 |
| The woman has high cheekbones. Her hair is black in colour with bangs. She has big lips with arched eyebrows and a slightly open mouth. The smiling, young attractive woman has rosy cheeks and heavy makeup. She's wearing earrings, necklace and lipstick. | 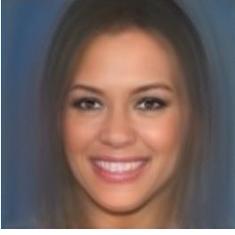 | The man has high cheekbones. He sports a 5 o'clock shadow. He has straight hair. He has big nose, narrow eyes with bushy eyebrows. The young attractive man is smiling. He's wearing necktie. | 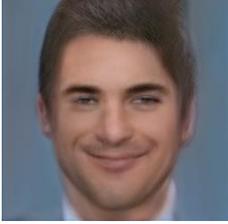 |
| The man has a chubby face. He sports a goatee and mustache. His hair is black in colour. He has big lips and big nose. The man looks young. He's wearing necktie. | 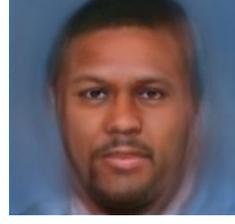 | The woman has high cheekbones. She has wavy hair which is blond in colour with bangs. She has pointy nose and a slightly open mouth. The smiling, young attractive woman has heavy makeup. She's wearing earrings and lipstick. | 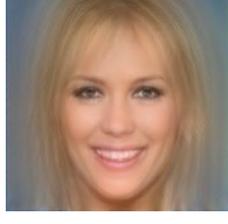 |
| The man has high cheekbones. He sports a goatee with sideburns. His hair is black in colour. He has big nose with bushy eyebrows and a slightly open mouth. The young attractive man is smiling. He's wearing necktie. | 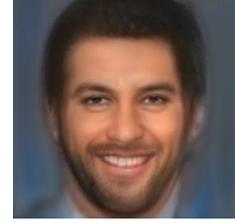 | The woman has high cheekbones. Her hair is brown in colour. She has arched eyebrows. The smiling, young attractive woman has rosy cheeks and heavy makeup. She's wearing lipstick. | 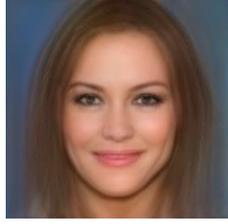 |
| The chubby double chined woman has high cheekbones. She has a receding hairline. She has big lips and big nose, narrow eyes with arched eyebrows and a slightly open mouth. The smiling, young woman has heavy makeup. She's wearing earrings, necklace and lipstick. | 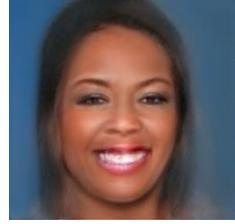 | The woman has oval face and high cheekbones. Her hair is black in colour with bangs. She has big lips and pointy nose with arched eyebrows and a slightly open mouth. The smiling, young attractive woman has rosy cheeks and heavy makeup. She's wearing earrings and lipstick. | 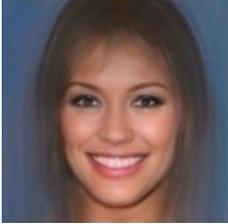 |

**Semantic alignment:** The second goal is to generate images that are consistent with a given description. Generating images that are close to the real images is the most natural way to show the semantic alignment of the images with their descriptions. Table 5 shows how close the generated images are to the real images corresponding to the description. Our model generates images that are similar to the real image to a certain extent. However, the generated images represent most of the features in the descriptions and it shows that the generated images have a good consistency with the descriptions.

Another interesting observation made during the generation of the images is the sensitivity to facial attributes in the descriptions. Table 6 shows how generated images through the proposed framework can be manipulated by changing the attributes in the description. Changes in the description are clearly shown on the generated images.  This shows how well the model has learnt the facial attributes in the descriptions and thereby, the semantic consistency of the generated images.

## 4.2. Quantitative Evaluation

The generated images are quantitatively evaluated for the quality and the semantic consistency using the FID score, and the similarity shown to the real image corresponding to the description using the Face Semantic Similarity and the Face Semantic Distance.

Table 4. Comparison with other face generation models using the FSD and FSS criterion.

| Model | FSD | FSS (%) |
|---|---|---|
| AttnGAN [11] | 1.269 | 59.28 |
| StackGAN [10] | 1.310 | - |
| FTGAN [23] | 1.267 | 59.41 |
| Realistic Image Generation of Face [24] | 1.118 | - |
| **Ours** | **0.9224** | **56.96** |

As shown in the above table, our model is performing very close to the state-of-the-artwork. From the table, we interpret that our model has an FSD of 0.9224 which is lower compared to AttnGAN[11], StackGAN[10], FTGAN[23], and Realistic Image Generation of Face from Text[24] models. This tells us that our model is capable of generating images closer to the ground truth better than the other models. Our model can generate images that are almost 57% closer to the ground truth images. This is most likely due to the limited dataset we have used in the training process. We believe if a larger dataset is used the results would be even better. However, these images show high consistency with the input descriptions. Our model generates images of resolution 1024x1024, whereas the other models are only capable of generating images of resolution 256x256. This is one of the biggest contributions of our model. The images generated with our model have an FID score of 118.097. However, we cannot directly compare the FID scores of these models to ours because of the difference in the resolution of the images.

## 5. CONCLUSION AND FUTURE WORK

Our principal objective was to develop a novel framework for T2F that can generate realistic, high-resolution images that are consistent with the input descriptions. The availability of high-resolution face generation models and the need to explore the use of them in T2F generation was the main motivation behind this paper. In that sense, we focused on three tasks.

- To utilize StyleGAN2 generator for T2F to produce high-resolution images.
- To generate images that are semantically aligned with the input descriptions.
- To measure the semantic consistency of the generated images against the real images.

We proposed a model, comprising the BERT language model, StyleGAN2 generator, and a text-to-latent space model to achieve those tasks. Here we embedded the descriptions in the latent space of the StyleGAN2 generator and thereby controlled the facial images generated using text descriptions.

With this approach, we have been able to achieve our goal of generating realistic facial images that are aligned with the input descriptions. Furthermore, we have made use of an existing high-resolution generator and opened up for more work on exploring the task of using the latent space of the StyleGAN2 for controlling the images generated using text descriptions. From both, the quantitative and qualitative evaluative comparisons we can see that the generated images exhibit good image quality and consistency with the input descriptions. We were able to generate images of 1024x1024 in resolution and these images showed 57% similarity to the real image, performing better than most recent work. We achieved these results using a smaller dataset, due to limitations in resources. Therefore, we can conclude that with a larger dataset and exposure to more facial attributes, this approach would produce even better results.

However, we still need to improve the image quality and the consistency with the descriptions. Appearance enhancing attributes like earrings, necklaces, caps were not visible in the images, and images generated using less frequent attributes did not show good realism. Therefore, we believe using a larger dataset would help tackle this issue when being exposed to more facial attributes. We further hope to work on focusing on the diversity of the facial images generated. So far in this model, we have only focused on generating one image per description. However, a single description corresponds to many facial images. Therefore, this too is an area that needs to be focused on.

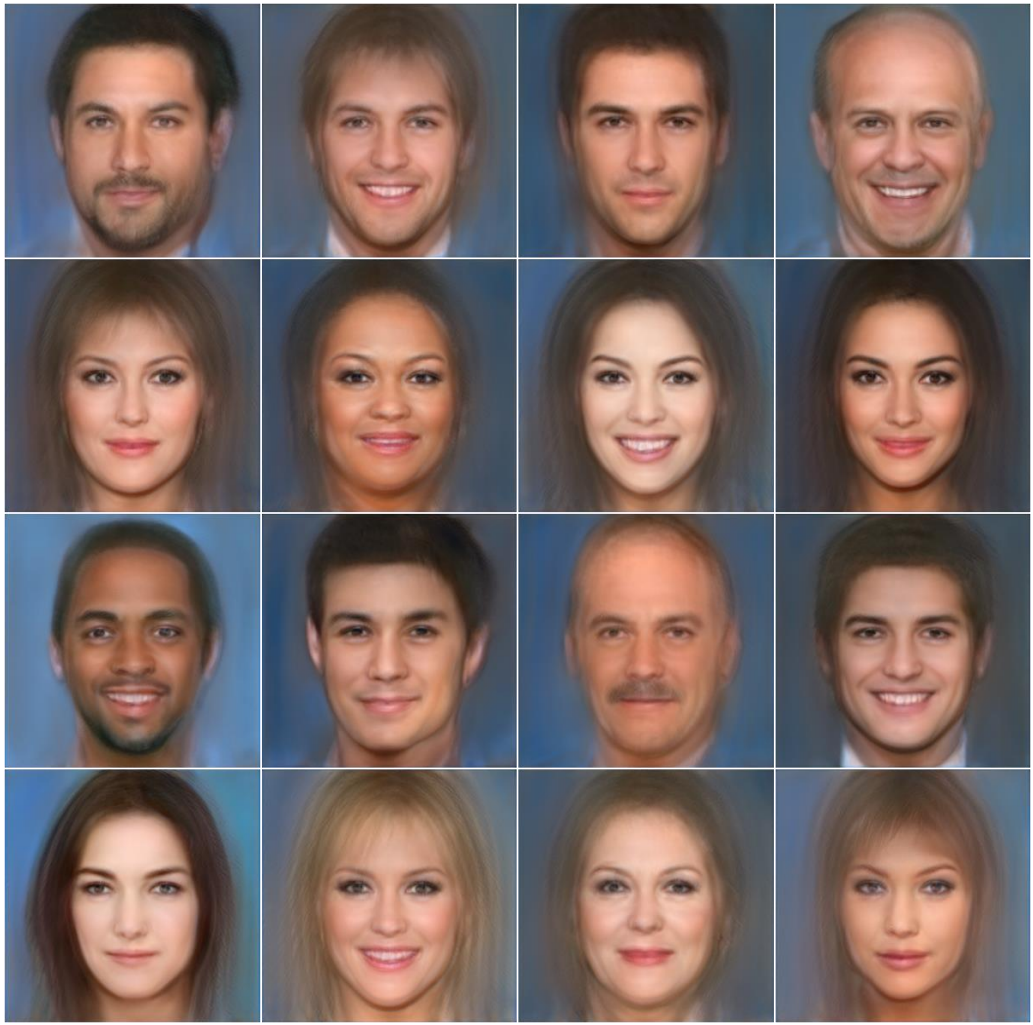

Figure 4. Sample images generated by the proposed model.

Table 5. Generated images with their corresponding real images and descriptions.

| Description | Real Image | Generated Image |
|---|---|---|
| The woman has oval face. Her hair is blond in colour. She has arched eyebrows. The smiling, young attractive woman has heavy makeup. She's wearing lipstick. | 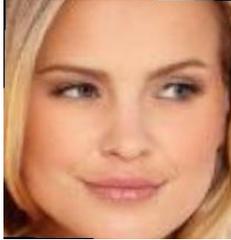 | 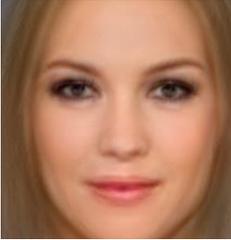 |
| The chubby double chined woman has high cheekbones. She has a receding hairline. She has big lips and big nose with arched eyebrows. The smiling, young woman has heavy makeup. She's wearing earrings and lipstick. | 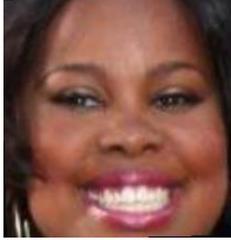 | 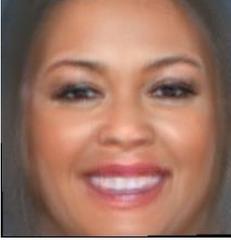 |
| The woman has oval face and high cheekbones. She has wavy hair. She has arched eyebrows and a slightly open mouth. The smiling, young attractive woman has rosy cheeks and heavy makeup. She's wearing earrings, necklace and lipstick. | 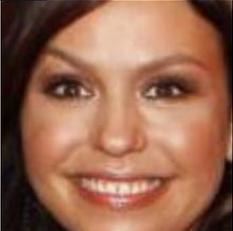 | 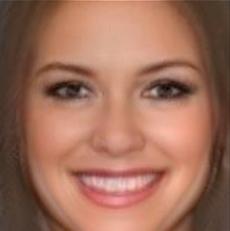 |
| The man has wavy hair which is black in colour. He has big lips and big nose with bushy eyebrows and a slightly open mouth. The young attractive man is smiling. He's wearing necktie. | 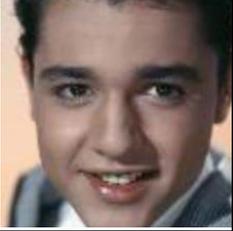 | 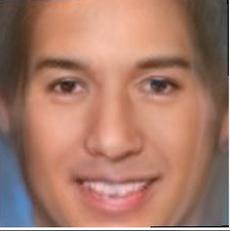 |
| The man has straight hair which is brown in colour. He has big nose. He's wearing necktie. | 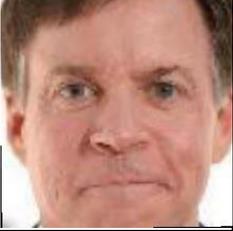 | 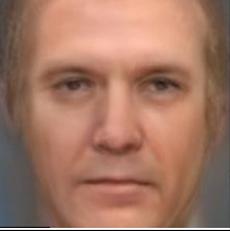 |
| The chubby double chined man has oval face and high cheekbones. He sports a 5 o'clock shadow, goatee and moustache. He is bald. He has big lips and big nose and a slightly open mouth. The man is smiling. He's wearing necktie. | 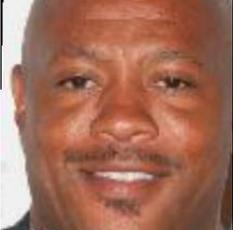 | 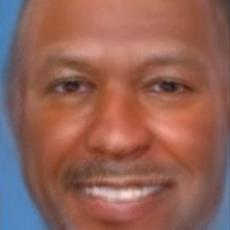 |

Table 6. Manipulating Generated Images.

The woman has high cheekbones. She has straight hair which is brown in colour with bangs. The smiling, young attractive woman has heavy makeup. She's wearing lipstick.

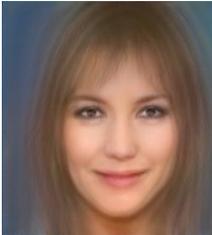

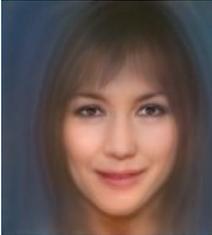 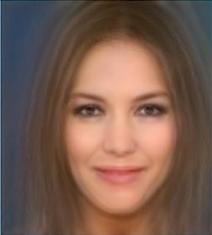 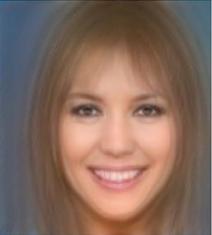 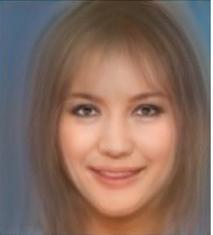 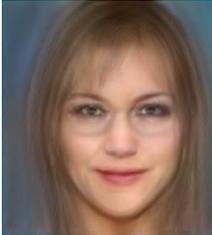

| Black Hair | No Bangs | Open Mouth | Chubby Face | Wearing Eyeglasses |
|---|---|---|---|---|

## AUTHORS

**Akila Ayanthi** is an undergraduate currently pursuing the BCS special degree from the University of Ruhuna, Sri Lanka. Her research interests include computer vision, image processing, deep learning and Generative Adversarial Networks (GANs).

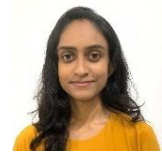

**Sarasi Munasinghe** received the PhD degree in computer science and engineering from the Queensland University of Technology, Australia in 2018. She completed her BSc. in Engineering from the University of Peradeniya, Sri Lanka in 2010. She currently works at the Department of Computer Science, University of Ruhuna as a senior lecturer. Her research interests are in the fields of deep learning, computer vision and image processing.

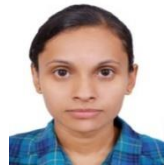